\documentclass[conference]{IEEEtran}
\usepackage{fancyhdr} 
\IEEEoverridecommandlockouts

\usepackage{cite}
\usepackage{amsmath,amssymb,amsfonts}
\usepackage{algorithmic}
\usepackage{graphicx}
\usepackage{textcomp}
\usepackage{xcolor}
\usepackage{booktabs}
\usepackage{multirow}
\usepackage{url}
\usepackage[hidelinks]{hyperref}

\usepackage{booktabs}
\usepackage{tabularx}
\usepackage{array}
\usepackage{makecell}

\def\BibTeX{{\rm B\kern-.05em{\sc i\kern-.025em b}\kern-.08em
    T\kern-.1667em\lower.7ex\hbox{E}\kern-.125emX}}

\title{Scalable In-Domain Self-Supervised Foundation Model for
Dense Representation Transfer in High-Resolution Plant Imaging}

\author{
\IEEEauthorblockN{
Junlin Guo,
Sharmin Majumder,
Isaac Lyngaas,
John Lagergren,
and Xiao Wang\textsuperscript{*}
}
\IEEEauthorblockA{
Oak Ridge National Laboratory\\
Oak Ridge, TN, USA\\
\{guoj3, majumders, lyngaasir, lagergrenjh, wangx2\}@ornl.gov
}
\thanks{This manuscript has been authored by UT-Battelle, LLC, under contract DE-AC05-00OR22725 with the US Department of Energy (DOE). The US government retains and the publisher, by accepting the article for publication, acknowledges that the US government retains a nonexclusive, paid-up, irrevocable, worldwide license to publish or reproduce the published form of this manuscript, or allow others to do so, for US government purposes. DOE will provide public access to these results of federally sponsored research in accordance with the DOE Public Access Plan ( https://www.energy.gov/doe-public-access-plan ).}
\thanks{\textsuperscript{*}Corresponding author: Xiao Wang
(wangx2@ornl.gov).}
}

\begin{document}

\maketitle

\thispagestyle{fancy} 
\lhead{} 
\rhead{} 
\chead{} 
\lfoot{\footnotesize{ 
SC26 Workshops, November 15-20, 2026, Chicago, Illinois, USA 
\newline 979-8-3195-1221-5/26/\$31.00 \copyright 2026 IEEE}} 
\rfoot{} 
\cfoot{} 
\renewcommand{\headrulewidth}{0pt} \renewcommand{\footrulewidth}{0pt}

% =========================
% Abstract
% =========================
\begin{abstract}
High-resolution plant imaging enables detailed characterization of plant
morphology, but dense scientific analysis remains limited by costly pixel-level
annotations, large image pixel dimensions, and substantial variation in imaging
conditions. This work proposes a scalable in-domain self-supervised pretrained
foundation model for high-resolution, high-pixel-dimension multi-species plant imagery.
A masked autoencoder with a ViT backbone is pretrained on more than 10 million
multi-view plant image tiles using distributed training. Following scalable
pretraining, the learned foundation-model representations are comprehensively
benchmarked across fine-grained dense prediction and coarse global feature
recognition, with particular emphasis on limited supervision and realistic
downstream imaging conditions. This work focuses on the domain gap of existing
foundation models in dense feature representation and transfer. Through
extensive experiments involving limited annotations, cross-view variation, and
resolution degradation, the in-domain FM achieves a Mean Dice of 0.8686 and a
Pooled Dice of 0.8959, outperforming an MAE counterpart pretrained on
large-scale natural-image data by 0.0694 and 0.0613, respectively. The results
further indicate that increasing pretraining scale produces consistent
improvements in dense feature transfer. Overall, these findings suggest that
scaling in-domain self-supervised pretraining can reduce the domain gap and
improve transferable dense representations for high-pixel-dimension scientific
imaging.
\end{abstract}


% =========================
% Main paper
% =========================
\section{Introduction}
% Write the introduction here. Cite papers using BibTeX keys, for example~\cite{example2025}.

Recent advances in high-resolution imaging and large-scale computing are rapidly increasing the volume, spatial detail, and diversity of data available for scientific discovery. Similar challenges arise across scientific imaging domains, such as earth observation~\cite{remote_sensing_bigdata}, computational pathology~\cite{guo2025evaluating} and spatial transcriptomic~\cite{asign}, where high-dimensional measurements require scalable computational analysis. In plant science, modern imaging systems enable fine-grained characterization of plant morphology and growth~\cite{plantphenotype1, plantphenotype2}. However, the ability to acquire increasingly detailed observations has outpaced our ability to extract dense scientific information from them at scale. Current deep learning (DL) and computer vision methods still rely heavily on labor-intensive pixel-level annotations~\cite{ahn2018affinity}. Transfer learning from models pretrained on natural images, such as ImageNet~\cite{imagenet}, can reduce this burden, but its effectiveness often degrades as the domain gap increases. This limitation becomes particularly important in scientific imaging, where models must contend with severe foreground-background imbalance, scale and resolution variation, and high-pixel-dimension images. Moreover, reliable analysis requires predictions to remain consistent not only at the local tile level, but also after reconstruction into the full-image space.

At the same time, plant imaging naturally produces large volumes of unlabeled, high-resolution, and multi-view observations, creating an opportunity to shift the learning paradigm from annotation-intensive supervised learning (SL) toward large-scale in-domain self-supervised representation learning (SSL). We hypothesize that scaling SSL pretraining over large scientific imaging datasets can produce reusable representations that reduce dependence on dense supervision and remain robust to practical imaging constraints, including limited labels, foreground-background imbalance, and scale and resolution variation.

\begin{figure*}[t]
    \centering
    \includegraphics[width=\textwidth]{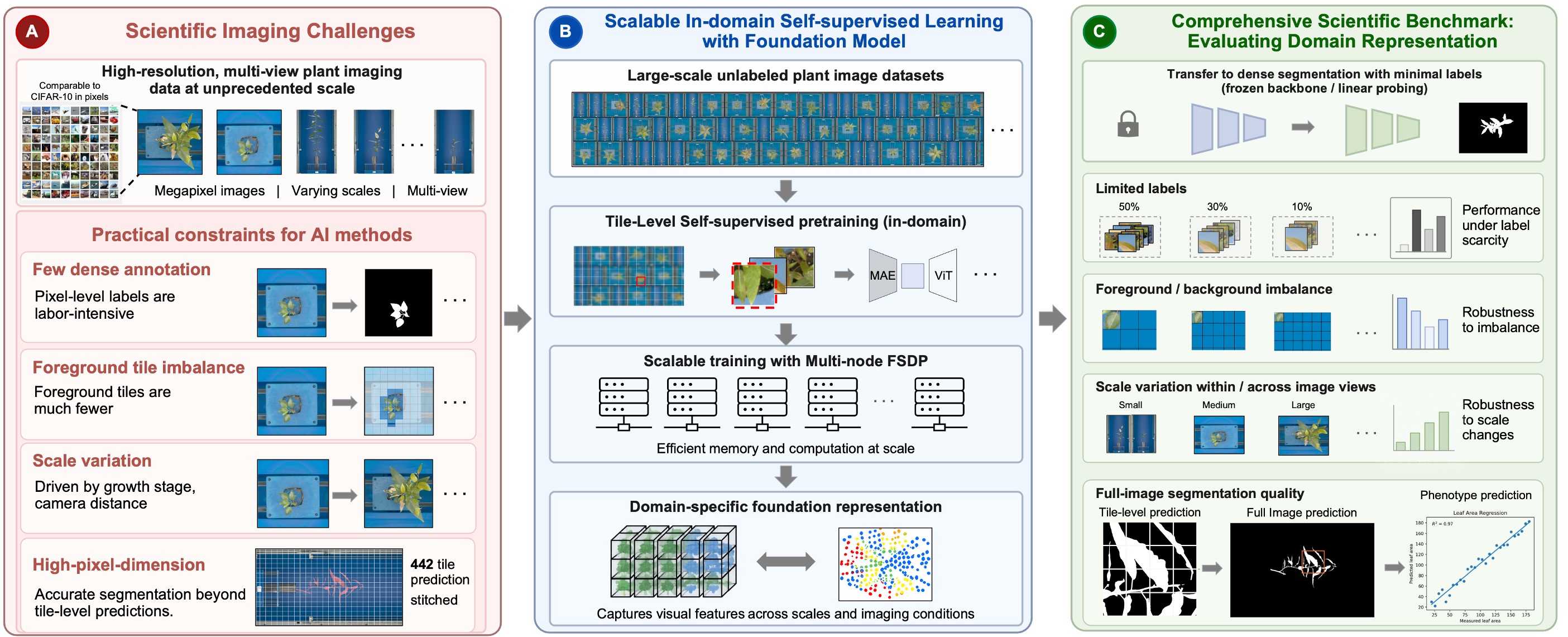}
    \caption{Overview of this work. (a) Practical scientific imaging challenges, including limited annotations,
foreground-background imbalance, scale and resolution variation, and
high-pixel-dimension imagery.
(b) Scalable self-supervised pretraining on large-scale polar plant image
tiles using MAE framework and distributed training. (c) Comprehensive evaluation of the learned representation under different realistic downstream conditions and phenotype quantification.}
    \label{fig:overview}
\end{figure*}

In this work, we investigate this hypothesis using large-scale plant imaging. An overview of the proposed framework is shown in Fig.~\ref{fig:overview}. First, Fig.~\ref{fig:overview}(a) summarizes the key challenges of dense scientific image analysis at scale, including limited fine-grained annotations, foreground-background imbalance, scale and resolution variation, and high-pixel-dimension images that require reliable full-image segmentation. Fig.~\ref{fig:overview}(b) presents our scalable learning framework, in which We pretrain a domain-specific self-supervised foundation model (FM) on high-resolution, multi-view, multi-species plant observations using distributed multi-node fully sharded data parallel (FSDP). Fig.~\ref{fig:overview}(c) illustrates our evaluation framework, where the learned representation is systematically benchmarked on downstream imaging tasks, particularly poplar, under practical imaging constraints using a frozen backbone with lightweight downstream adaptation. Through extensive experiments, we evaluate the generalizability of the learned representation under these imaging constraints, with all dense prediction tasks assessed after reconstructing tile-level predictions into the full-image space. We further examine the scientific utility of the resulting full-image segmentations through downstream phenotype estimation, using leaf-area prediction as an example. The code for the comprehensive benchmarking will be publicly available at \url{https://github.com/junlinguo/PlantVFM-DenseBench}.

The contributions of this work are threefold:
\begin{itemize}
    \item This work proposes a scalable in-domain self-supervised learning
    framework for high-resolution, multi-view, multi-species plant imaging, with distributed MAE pretraining over more than 10 million image tiles.

    \item A comprehensive evaluation is conducted for dense feature representation
transfer under practical scientific imaging constraints, including limited
annotations, cross-view variation, resolution degradation, and tile-to-full-image
prediction aggregation.

\item Increasing the scale of in-domain pretraining consistently improves
dense feature transfer for plant scientific imaging, outperforming
natural-image pretrained representations.
\end{itemize}

\section{Related Work}

\subsection{Self-Supervised Learning and Vision Foundation Models}

Self-supervised learning has become a widely adopted paradigm for
learning visual representations from large-scale unlabeled data. Existing
approaches are commonly built around representation-matching and predictive
reconstruction objectives. Contrastive and self-distillation methods, such as
SimCLR~\cite{simclr}, MoCo~\cite{moco}, and DINO~\cite{dino}, learn representations by enforcing consistency across
different views of an image. In contrast, masked image
modeling approaches such as BEiT~\cite{beit} and masked autoencoders (MAE)~\cite{mae} learn features
by predicting or reconstructing masked image content. Such
masked prediction objectives are particularly relevant to dense representation
learning; For example, DINOv2 and DINOv3 incorporate an iBOT-style masked-image
modeling objective to enhance patch-level representations, which has been
shown to benefit dense prediction performance~\cite{ibot,dinov2,dinov3}.
Recent vision foundation models increasingly combine these complementary SSL
objectives, demonstrating the effectiveness of scalable pretraining for both
global and spatially localized visual representations.

\subsection{Domain-Specific Foundation Models for Scientific Imaging}

The domain gap between natural and scientific imagery has motivated the development of foundation models pretrained directly on domain-specific datasets~\cite{lu2025vision, vorontsov2023virchow, szwarcman2025prithvi, cong2022satmae}. In computational pathology, Virchow~\cite{vorontsov2023virchow} scales self-supervised learning to large collections of whole-slide images and demonstrates transferable representations across different pathological downstream tasks. In Earth observation,
Prithvi leverages large-scale satellite imagery for geospatial representation learning~\cite{szwarcman2025prithvi}, while SatMAE extends masked autoencoding to temporal and multispectral observations~\cite{cong2022satmae}. Scale-MAE further incorporates spatial scale into masked
pretraining to improve multiscale remote-sensing representations~\cite{reed2023scale}. DeepAndes develops a self-supervised vision foundation model specifically for multi-spectral remote-sensing imagery of the Andean archaeology~\cite{11196959}. At larger computational scales, ORBIT-2 demonstrates exascale vision foundation model training for weather and climate downscaling~\cite{wang2025orbit}. Together, these studies demonstrate the value of adapting
large-scale self-supervised pretraining to scientific data characteristics,
while dense representation transfer under limited annotations and substantial
resolution variation remains less systematically explored.

\section{Method}
\label{sec:method}

\begin{figure*}[t]
    \centering
    \includegraphics[width=\textwidth]{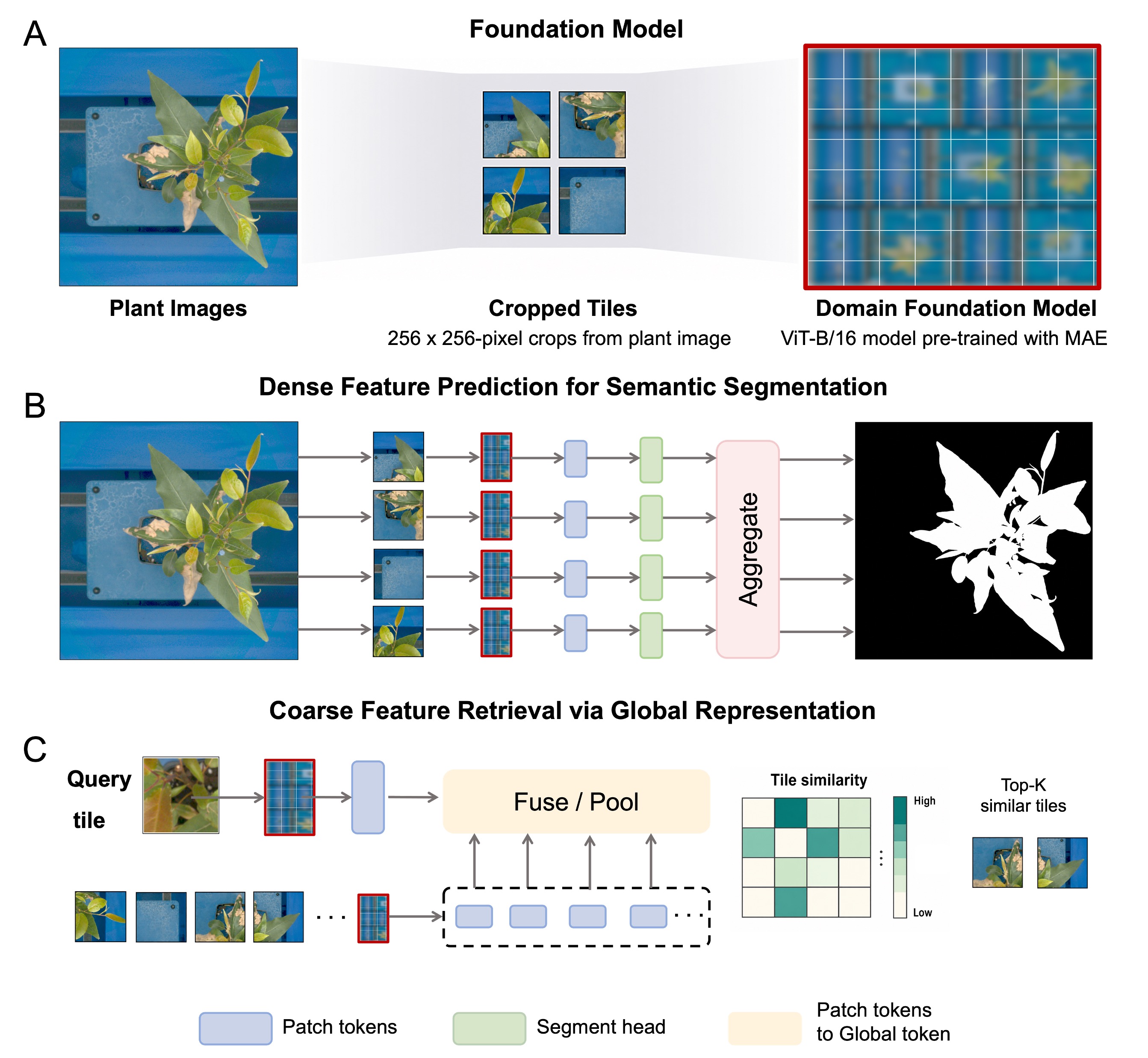}
\caption{Feature representation and downstream transfer framework.
(a) Large-scale self-supervised pretraining on multi-species plant image tiles using a
masked autoencoder (MAE) with a ViT backbone. (b) For dense feature prediction,
spatial patch features are passed to a lightweight MLP segmentation head with
the backbone frozen, followed by full-image prediction aggregation.
(c) For coarse feature retrieval, patch-token embeddings are fused using mean
or max pooling to form global representations for zero-shot tile-to-tile
retrieval.}
    \label{fig:feature_transfer}
\end{figure*}

\subsection{Large-Scale Scientific Imaging Dataset and Preprocessing}

We use two RGB imaging views of multi-species plants: RGB1 side-view images acquired at approximately $0.25~\mathrm{mm/px}$ and RGB2 top-down images acquired at approximately $0.0167~\mathrm{mm/px}$. The dataset contains more than 160,000 high-pixel-dimension images, with individual images reaching approximately 7,000 pixels along each spatial dimension. For foundation-model pretraining, each image is partitioned into $256\times256$ tiles with 50\% overlap, yielding more than 10 million valid training tiles after excluding purely dark-background tiles with zero pixel-intensity variance. To account for differences in illumination and acquisition conditions across imaging views, each tile is independently min--max normalized before being distributed across GPU ranks for FSDP training.

\subsection{Scalable In-Domain Self-Supervised Pretraining}
Coarse-grained plant recognition often transfers well from natural-image pretraining, whereas dense prediction requires spatially precise representations under limited pixel-level supervision. We therefore adopt masked auto-encoding to learn dense, in-domain representations from large-scale unlabeled plant images.

\paragraph{Masked-Autoencoder Pretraining} As shown in Fig.~\ref{fig:feature_transfer}(a), we use the MAE as the self-supervised pretraining framework for a ViT-B/16 encoder. Each $256\times256$ tile is divided into $16\times16$ patches, producing 256 patch tokens with an embedding dimension of 768. The 12-layer ViT-B encoder is pretrained with a 50\% random masking ratio, while an asymmetric decoder reconstructs the masked patches using a mean-squared error (MSE) loss. The pretrained encoder serves as the domain-specific representation backbone for all downstream evaluations.

\paragraph{Distributed Training at Scale}
To scale in-domain pretraining over the large scientific imaging dataset, we use FSDP across multiple compute nodes. Model computation and data are distributed across multiple accelerators, with activation checkpointing used to reduce memory overhead. This setup enables scalable MAE pretraining and supports the same training workflow for larger model configurations.

\subsection{Feature Representation and Downstream Transfer}

As shown in Fig.~\ref{fig:feature_transfer}(a), the pretrained ViT backbone
produces patch-level embeddings that serve as shared representations for
downstream evaluation. We evaluate these features through dense semantic
segmentation and complementary coarse-grained retrieval, as illustrated in
Fig.~\ref{fig:feature_transfer}(b) and (c).

\paragraph{Dense Feature Prediction}
Localized patch tokens are preserved and passed to a lightweight
multilayer perceptron (MLP) segmentation head. The pretrained backbone remains
frozen, and tile-level predictions are reconstructed into the original
full-image space for evaluation.

\paragraph{Coarse Feature Retrieval}
Patch-token embeddings are fused using mean or max pooling to form a global
tile representation. Without task-specific fine-tuning, these features are
used for zero-shot tile-to-tile retrieval based on the embedding similarity.

\subsection{Full-Image Dense Representation Benchmark}

The primary evaluation focuses on dense semantic prediction under realistic
scientific imaging constraints. Throughout the benchmark, the pretrained FM
backbone remains frozen and only a lightweight segmentation head is adapted,
allowing us to assess the transferability of the learned in-domain
representation without full-model fine-tuning. A subset of 55
high-pixel-dimension plant images spanning side and top-down views is manually
annotated for semantic segmentation evaluation.

\paragraph{Full-Image Segmentation Evaluation} Although inference is performed at the tile level, all predictions are stitched back to their original spatial locations and evaluated in the full-image space. This protocol reflects scientific applications that require spatially consistent segmentation across the complete image.

\paragraph{Limited Annotation Setting}
To evaluate label efficiency, we train the lightweight segmentation head using 100\%, 50\%, 30\%, and 10\% of the available pixel-level annotations. All settings use the same full-image reconstruction and evaluation protocol. This experiment quantifies how effectively each pretrained representation supports dense prediction as supervision becomes increasingly limited.

\paragraph{Scale and Resolution Variation} We evaluate dense representation transfer under two complementary imaging variations. First, we compare RGB1 side-view and RGB2 top-down images, which differ in viewpoint, apparent plant scale, and acquisition resolution. Second, we perform controlled resolution degradation on RGB1 by progressively downsampling the original images while keeping the imaging modality and full-image evaluation protocol unchanged. Together, these settings assess representation transfer across naturally varying acquisition conditions and reduced image resolution.

\paragraph{Foreground--Background Training Distribution} High-resolution plant images naturally exhibit strong foreground-background imbalance. We use a foreground-to-background tile ratio of 1:3 as the baseline downstream training distribution and further evaluate a background-enriched setting using all available background tiles (approximately 1:15). The pretrained backbone remains frozen, and evaluation is performed on the same naturally imbalanced full-image test set. This setting assesses the sensitivity of dense prediction to foreground-background composition during downstream training.

\subsection{Phenotype Quantification}
As a downstream extension motivated by plant-science applications, we derive quantitative phenotypes from the reconstructed full-image segmentations. Using 2D projected leaf area as a representative example, phenotype values are estimated directly from the predicted masks and compared with manual reference measurements through regression analysis. This evaluation examines whether representations trained at scale, when coupled with a lightweight MLP segmentation head, can support reliable domain-specific scientific measurements with reduced downstream model complexity and reliance on dense annotations.

\section{Data and Experimental Setup}
\label{sec:experiments}

\subsection{Pretraining and Downstream Data}
\label{sec:data}

\paragraph{Pretraining Data}
We use the large-scale multi-view multi-species plant imaging dataset described in
Sec.~\ref{sec:method} for self-supervised pretraining. The dataset contains
more than 160,000 RGB images and over 10 million valid $256\times256$ training tiles after preprocessing. In this work, SSL pretraining uses unlabeled multi-species plant images, whereas the downstream benchmark focuses on poplar; some downstream images are included in SSL pretraining but are used without labels.

\paragraph{Downstream Segmentation Data}
For dense representation evaluation, we use 55 annotated high-pixel-dimension plant images spanning RGB1 side-view and RGB2 top-down observations. The two views differ substantially in spatial resolution, with RGB1 acquired at approximately $0.25~\mathrm{mm/px}$ and RGB2 at approximately $0.0167~\mathrm{mm/px}$. RGB1 provides the primary downstream benchmark and represents a more challenging transfer setting due to its coarser spatial sampling and distinct imaging characteristics relative to natural-image appearance, while RGB2 is additionally used to evaluate representation transfer across view and acquisition scale. \textcolor{black}{Details of the annotation procedure used to generate the ground-truth segmentation masks are provided in~\cite{milligan2026vision}}.

Each whole-plant image can be cropped into approximately 200 non-overlapping $256\times256$ tiles. The training, validation, and test sets are split at the plant-image level using a \(7{:}1{:}2\) ratio to prevent tile leakage across splits. Segmentation metrics are computed at the plant-image level. Despite the substantial number of evaluation tiles, the relatively small number of annotated plant images remains a limitation of the benchmark.

\subsection{Baseline Models}
\label{sec:baselines}
We compare the proposed in-domain FM representation with three baselines: a
randomly initialized ViT, ImageNet-pretrained MoCo-v2~\cite{moco}, and ImageNet-pretrained
MAE~\cite{mae}. For representation evaluation, pretrained backbones remain frozen and
only the lightweight downstream segmentation head is optimized. All methods
use the same downstream data split and full-image evaluation protocol.

\subsection{Dense Prediction Evaluation}
\label{sec:dense_eval}

% \paragraph{Main Benchmark}
% The primary benchmark evaluates full-image semantic segmentation using the
% complete downstream annotation set. Tile-level predictions are reconstructed
% into the original image space before evaluation.

\paragraph{Main Benchmark}
The primary benchmark evaluates full-image semantic segmentation using
100\% of the available downstream training annotations. Tile-level
predictions are reconstructed into the original image space before evaluation.

\paragraph{Limited Annotation}
Label efficiency is evaluated using 100\%, 50\%, 30\%, and 10\% of the
available downstream annotations while keeping the evaluation set fixed.

\paragraph{Cross-View Evaluation}
We separately evaluate RGB1 side-view and RGB2 top-down images to examine representation transfer across different viewpoints, spatial scales, and acquisition resolutions.

\paragraph{Resolution Degradation}
To evaluate sensitivity to reduced image resolution, RGB1 images are
downsampled by $5\times$ while keeping the remaining evaluation protocol
unchanged.

\paragraph{Downstream Training Sensitivity}
We further evaluate the effect of foreground-background training composition
and segmentation-loss weighting. The standard foreground-to-background tile
ratio of 1:3 is compared with a background-enriched setting using all
available background tiles, approximately 1:15.

\subsection{Coarse Feature Retrieval}
\label{sec:retrieval}

To evaluate global tile-level feature representations, patch-token embeddings
are fused using mean or max pooling to form a single global representation.
Without task-specific fine-tuning, we perform zero-shot tile-to-tile retrieval
based on embedding similarity.

\subsection{Pretraining Scale Evaluation}
\label{sec:scaling}

% \paragraph{Pretraining Scale Evaluation}

To evaluate the effect of increasing pretraining scale, we compare FMs
obtained at 25\%, 50\%, and 100\% of the cumulative pretraining data exposure.
Each FM is evaluated using the same primary dense segmentation benchmark
with a frozen backbone and identical downstream training and evaluation
methods.

\subsection{Evaluation Metrics}
\label{sec:metrics}

\paragraph{Dense Segmentation}
We report Mean Dice and Pooled Dice to evaluate segmentation performance in
the reconstructed full-image space. For a high-pixel-dimension image $i$, the
Dice score is defined as
\begin{equation}
    \mathrm{Dice}_i =
    \frac{2TP_i}{2TP_i + FP_i + FN_i},
\end{equation}
where $TP_i$, $FP_i$, and $FN_i$ denote the pixel-level true positives,
false positives, and false negatives, respectively. Mean Dice averages the
Dice score across all $N$ evaluation images,
\begin{equation}
    \mathrm{Mean~Dice} =
    \frac{1}{N}\sum_{i=1}^{N}\mathrm{Dice}_i,
\end{equation}
while Pooled Dice aggregates pixel-level predictions over the complete
evaluation set before computing Dice,
\begin{equation}
    \mathrm{Pooled~Dice} =
    \frac{2\sum_i TP_i}
    {2\sum_i TP_i+\sum_i FP_i+\sum_i FN_i}.
\end{equation}
Together, these metrics characterize both image-level consistency and overall
segmentation performance across the evaluation set.

\paragraph{Tile-to-Tile Image Retrieval}
For coarse representation evaluation, tiles are ranked according to cosine
similarity between their global feature embeddings. For query $q$, average
precision within the top-$K$ retrieved tiles is computed as
\begin{equation}
    AP_q@K =
    \frac{1}{\min(R_q,K)}
    \sum_{j=1}^{K} P_q(j)\,\mathrm{rel}_q(j),
\end{equation}
where $R_q$ denotes the number of relevant tiles, $P_q(j)$ is the precision
among the top-$j$ retrieved tiles, and $\mathrm{rel}_q(j)$ indicates whether
the tile at rank $j$ is relevant. Mean average precision is then computed
across all $Q$ queries as
\begin{equation}
    \mathrm{mAP}@K =
    \frac{1}{Q}\sum_{q=1}^{Q} AP_q@K.
\end{equation}
This work uses mAP@$K$ for $K=\{5,20,50,100\}$ to evaluate retrieval quality across
different neighborhood sizes.

\section{Results}

\subsection{Main Full-Image Dense Representation Benchmark}
\label{sec:main_results}

Table~\ref{tab:main_dense} compares frozen representations for full-image
plant segmentation. The in-domain FM achieves the best performance, with a
Mean Dice of 0.8686 and a Pooled Dice of 0.8959. Compared with ImageNet MAE,
the strongest natural-image pretrained baseline, this represents absolute
improvements of 0.0694 and 0.0613, respectively. ImageNet MoCo-v2 performs
substantially worse, while the randomly initialized frozen backbone provides
little useful dense representation. These results highlight the limitation of
generic natural-image pretraining for dense scientific prediction and show
that large-scale in-domain self-supervised pretraining produces substantially
more transferable spatial representations.

\begin{table}[h]
    \centering
    \caption{Full-image dense segmentation performance under the standard downstream setting. Best metric values are shown in
\textbf{bold}.}
    \label{tab:main_dense}

    \scriptsize
    \renewcommand{\arraystretch}{1.1}
    \setlength{\tabcolsep}{2.2pt}

    \begin{tabularx}{\columnwidth}{
        @{}
        >{\raggedright\arraybackslash}X
        >{\centering\arraybackslash}p{0.12\columnwidth}
        >{\centering\arraybackslash}p{0.22\columnwidth}
        >{\centering\arraybackslash}p{0.13\columnwidth}
        >{\centering\arraybackslash}p{0.14\columnwidth}
        @{}
    }
        \toprule
        \textbf{Model} &
        \textbf{Backbone} &
        \makecell{\textbf{Pretraining}\\\textbf{Domain}} &
        \makecell{\textbf{Mean}\\\textbf{Dice}} &
        \makecell{\textbf{Pooled}\\\textbf{Dice}} \\
        \midrule

        Random &
        ViT &
        None &
        0.0363 &
        0.0366 \\

        ImageNet MoCo-v2 &
        CNN &
        Natural images &
        0.4865 &
        0.5416 \\

        ImageNet MAE &
        ViT &
        Natural images &
        0.7992 &
        0.8346 \\

        \textbf{OPAL} &
        ViT &
        \textbf{Plant images} &
        \textbf{0.8686} &
        \textbf{0.8959} \\

        \bottomrule
    \end{tabularx}
\end{table}

\subsection{Label-Efficient Dense Transfer}

Table~\ref{tab:limited_annotations} evaluates dense representation transfer under limited downstream annotation settings. The in-domain FM consistently outperforms both natural-image pretrained baselines across all annotation budgets. With only 10\% of the annotations, it retains a Mean Dice of 0.7494 and a Pooled Dice of 0.7960, compared with 0.5989 and 0.6031 for ImageNet MAE. Performance degrades only gradually as supervision is reduced from 100\% to 50\% and 30\%, indicating that the domain SSL pretrained FM requires less pixel-level supervision for effective dense feature transfer.

\begin{table}[h]
    \centering
    \caption{Full-image dense segmentation performance under limited annotations. Best metric values are shown in
\textbf{bold}.}
    \label{tab:limited_annotations}

    \setlength{\tabcolsep}{3.0pt}
    \renewcommand{\arraystretch}{1.1}

    \resizebox{\columnwidth}{!}{
    \begin{tabular}{lcccccccc}
        \toprule
        \multirow{2}{*}[-1ex]{\textbf{Model}} &
        \multicolumn{2}{c}{\textbf{100\%}} &
        \multicolumn{2}{c}{\textbf{50\%}} &
        \multicolumn{2}{c}{\textbf{30\%}} &
        \multicolumn{2}{c}{\textbf{10\%}} \\
        \cmidrule(lr){2-3}
        \cmidrule(lr){4-5}
        \cmidrule(lr){6-7}
        \cmidrule(lr){8-9}

        &
        \textbf{Mean} & \textbf{Pooled} &
        \textbf{Mean} & \textbf{Pooled} &
        \textbf{Mean} & \textbf{Pooled} &
        \textbf{Mean} & \textbf{Pooled} \\
        \midrule

        ImageNet MoCo-v2 &
        0.4865 & 0.5416 &
        0.3509 & 0.3563 &
        0.1366 & 0.1322 &
        0.0129 & 0.0132 \\

        ImageNet MAE &
        0.7992 & 0.8346 &
        0.7671 & 0.7911 &
        0.7229 & 0.7447 &
        0.5989 & 0.6031 \\

        \textbf{OPAL} &
        \textbf{0.8686} & \textbf{0.8959} &
        \textbf{0.8543} & \textbf{0.8834} &
        \textbf{0.8247} & \textbf{0.8608} &
        \textbf{0.7494} & \textbf{0.7960} \\

        \bottomrule
    \end{tabular}
    }
\end{table}

\subsection{Scale and Resolution Variation}

\paragraph{Cross-View Variation}
Table~\ref{tab:cross_view} compares dense representation transfer across RGB1
side-view and RGB2 top-down imagery. All pretrained models achieve higher
segmentation performance on RGB2, while the in-domain FM consistently
outperforms the natural-image pretrained baselines across both views. On RGB2,
the in-domain FM reaches a Mean Dice of 0.9057 and a Pooled Dice of 0.9511,
compared with 0.8738 and 0.9017 for ImageNet MAE. Together with its stronger
performance on RGB1, these results demonstrate that the learned FM
representation through large-scale SSL pretraining transfers effectively across substantially different imaging
views and acquisition conditions.

\begin{table}[h]
    \centering
    \caption{Full-image segmentation performance across RGB1 side-view and
    higher-resolution RGB2 top-down imagery. Best metric values are shown in
\textbf{bold}.}
    \label{tab:cross_view}

    \small
    \setlength{\tabcolsep}{3.2pt}
    \renewcommand{\arraystretch}{1.1}

    \begin{tabular}{lcccc}
        \toprule
        \multirow{2}{*}[-1ex]{\textbf{Model}} &
        \multicolumn{2}{c}{\textbf{RGB1 (Side View)}} &
        \multicolumn{2}{c}{\textbf{RGB2 (Top-Down)}} \\
        \cmidrule(lr){2-3}
        \cmidrule(lr){4-5}

        &
        \textbf{Mean} &
        \textbf{Pooled} &
        \textbf{Mean} &
        \textbf{Pooled} \\
        \midrule

        Random &
        0.0363 & 0.0366 &
        0.0882 & 0.1254 \\

        ImageNet MoCo-v2 &
        0.4865 & 0.5416 &
        0.5964 & 0.6905 \\

        ImageNet MAE &
        0.7992 & 0.8346 &
        0.8738 & 0.9017 \\

        \textbf{OPAL} &
        \textbf{0.8686} & \textbf{0.8959} &
        \textbf{0.9057} & \textbf{0.9511} \\

        \bottomrule
    \end{tabular}
\end{table}

\paragraph{Resolution Degradation}
Table~\ref{tab:resolution_degradation} evaluates dense segmentation under a
controlled $5\times$ resolution reduction on RGB1 imagery. Performance
decreases for all pretrained models after downsampling, showing the
sensitivity of fine-grained vision task to reduced spatial detail. Nevertheless, the
in-domain FM shows a Mean Dice of 0.6332 and a Pooled Dice of 0.6866, compared with 0.5736 and 0.6100 for ImageNet MAE. These results indicate that in-domain pretraining provides more
robust dense representations under substantial resolution degradation.

\begin{table}[h]
    \centering
    \caption{Full-image segmentation performance on RGB1 under controlled
    resolution degradation.  Best and second-best metric values are
shown in \textbf{bold} and \underline{underlined}, respectively.}
    \label{tab:resolution_degradation}
    \scriptsize
    \setlength{\tabcolsep}{3.5pt}
    \renewcommand{\arraystretch}{1.1}
    \resizebox{\columnwidth}{!}{
    \begin{tabular}{lcccc}
        \toprule
        & \multicolumn{2}{c}{\textbf{Original Resolution}}
        & \multicolumn{2}{c}{\textbf{5$\times$ Downsampled}} \\
        \cmidrule(lr){2-3}
        \cmidrule(lr){4-5}
        \multirow{2}{*}[+3.5ex]{\textbf{Model}}
        & \textbf{Mean}
        & \textbf{Pooled}
        & \textbf{Mean}
        & \textbf{Pooled} \\
        \midrule
        ImageNet MoCo-v2
        & 0.4865 & 0.5416
        & 0.2928 & 0.3026 \\

        ImageNet MAE
        & \underline{0.7992} & \underline{0.8346}
        & \underline{0.5736} & \underline{0.6100} \\

        In-domain FM (MAE)
        & \textbf{0.8686} & \textbf{0.8959}
        & \textbf{0.6332} & \textbf{0.6866} \\
        \bottomrule
    \end{tabular}
    }
\end{table}

\begin{figure*}[!t]
    \centering
    \includegraphics[width=\textwidth]{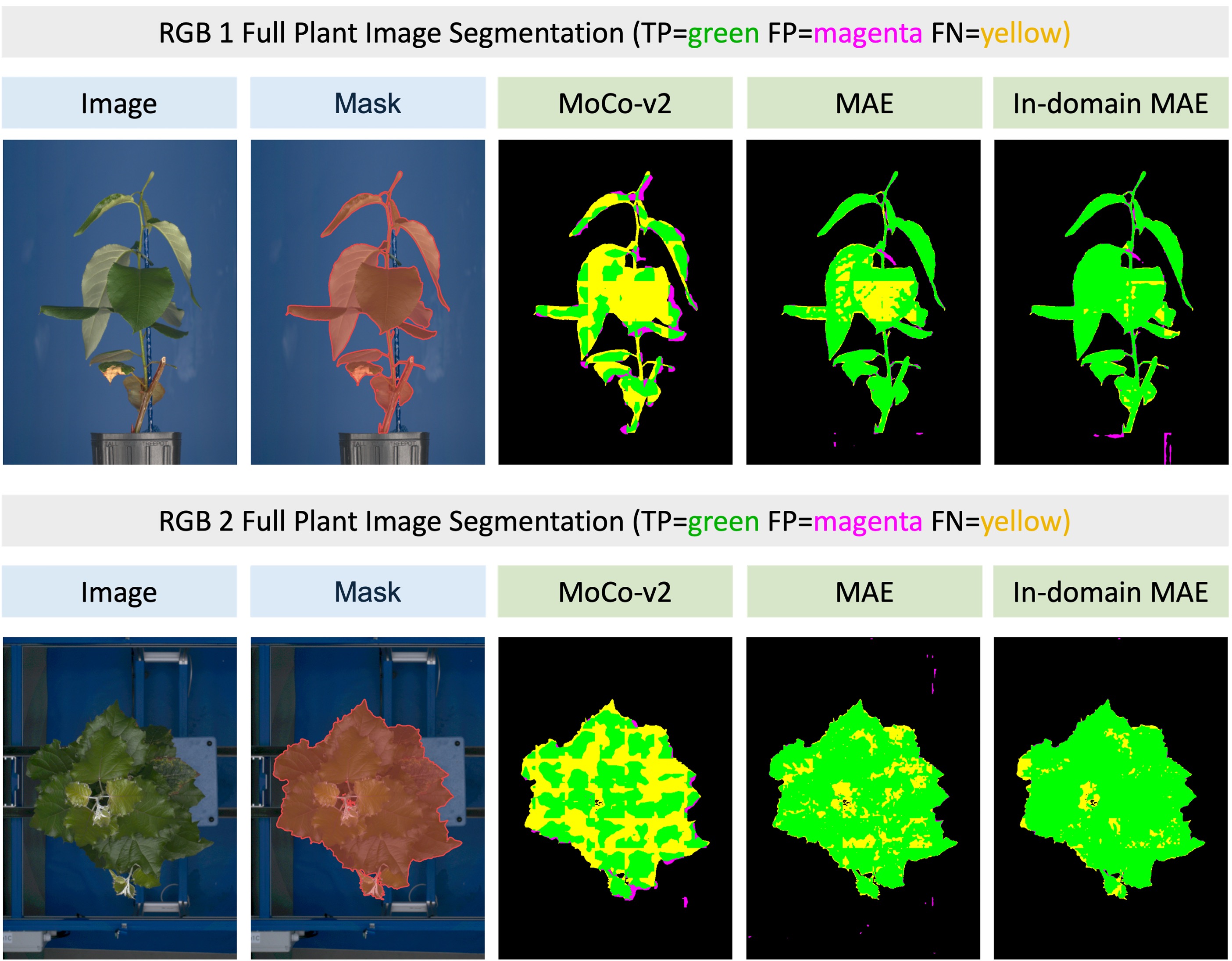}
    \caption{Qualitative Segmentation Performance for RGB1 (Side-View) and RGB2 (Top-Down) poplar whole-plant images. True-positive (TP),
    false-positive (FP), and false-negative (FN) regions are shown in green,
    magenta, and yellow, respectively.}
    \label{fig:qualitative}
\end{figure*}

\paragraph{Qualitative Evaluation}
Fig.~\ref{fig:qualitative} provides representative full-image segmentation
results across RGB1 and RGB2 imagery. The in-domain FM produces more complete
foreground predictions and reduces several false-positive and false-negative
regions relative to the natural-image pretrained baselines, consistent with
the quantitative results.

\subsection{Pretraining Data Scaling}

\begin{table}[h]
    \centering
    \caption{Dense segmentation performance across increasing pretraining data exposure at the 25\%, 50\%, and 100\% pretraining stages. Best metric values are shown in \textbf{bold}.}
    \label{tab:pretraining_scale}
    \scriptsize
    \setlength{\tabcolsep}{6pt}
    \renewcommand{\arraystretch}{2}
    \begin{tabular}{ccc}
        \toprule
        \textbf{Pretraining Scale}
        & \textbf{Mean Dice}
        & \textbf{Pooled Dice} \\
        \midrule
        25\%  & 0.8121 & 0.8398 \\
        50\%  & 0.8432 & 0.8671 \\
        100\% & \textbf{0.8686} & \textbf{0.8959} \\
        \bottomrule
    \end{tabular}
\end{table}

Table~\ref{tab:pretraining_scale} examines the FM performance as the pretraining scale increases. Segmentation
performance improves consistently from 25\% to 100\% pretrained data exposure, with Mean Dice
increasing from 0.8121 to 0.8686 and Pooled Dice from 0.8398 to 0.8959.
The intermediate 50\% model achieves 0.8432 Mean Dice and 0.8671 Pooled Dice.
Although the gains are gradual, the monotonic trend suggests further benefits
from scaling pretraining data exposure and compute, and motivates exploring
larger model capacities to better utilize this increased scale.

\subsection{Coarse Feature Evaluation}
\label{sec:coarse_results}

Table~\ref{tab:coarse_retrieval} evaluates zero-shot tile-to-tile retrieval
using global representations from patch-token fusion. Even the randomly
initialized model achieves relatively high performance, suggesting that this
coarse global-level task is comparatively easy and less sensitive to the
natural-to-plant domain gap. ImageNet MoCo-v2 performs best with mean pooling,
while the in-domain FM performs best with max pooling across all $K$. Overall,
domain-specific pretraining provides only a modest benefit for coarse retrieval
on this dataset compared with its stronger effect on dense spatial prediction.

\begin{table}[h]
    \centering
    \caption{Coarse feature tile-to-tile retrieval performance under different patch-token fusion strategies. For each fusion strategy, best and second-best metric values are
shown in \textbf{bold} and \underline{underlined}, respectively.}
    \label{tab:coarse_retrieval}

    \setlength{\tabcolsep}{2.6pt}
    \renewcommand{\arraystretch}{1.5}

    \resizebox{\columnwidth}{!}{
    \begin{tabular}{llcccc}
        \toprule
        \makecell{\textbf{Fusion}\\\textbf{Strategy}} &
        \textbf{Model} &
        \textbf{mAP@5} &
        \textbf{mAP@20} &
        \textbf{mAP@50} &
        \textbf{mAP@100} \\
        \midrule

        \multirow{4}{*}{Mean-Pool}
        & Random
        & 0.8616 & 0.8170 & 0.7819 & 0.7582 \\

        & ImageNet MoCo-v2
        & \textbf{0.9603}
        & \textbf{0.9456}
        & \textbf{0.9319}
        & \textbf{0.9175} \\

        & ImageNet MAE
        & 0.9513 & 0.9256 & 0.9038 & 0.8843 \\

        & In-domain FM (MAE)
        & \underline{0.9567}
        & \underline{0.9383}
        & \underline{0.9199}
        & \underline{0.9017} \\
        \midrule

        \multirow{4}{*}{Max-Pool}
        & Random
        & 0.9518 & 0.9317 & 0.9061 & 0.8796 \\

        & ImageNet MoCo-v2
        & \underline{0.9749}
        & \underline{0.9676}
        & \underline{0.9605}
        & \underline{0.9518} \\

        & ImageNet MAE
        & 0.9689 & 0.9529 & 0.9360 & 0.9183 \\

        & In-domain FM (MAE)
        & \textbf{0.9830}
        & \textbf{0.9753}
        & \textbf{0.9704}
        & \textbf{0.9646} \\

        \bottomrule
    \end{tabular}
    }
\end{table}

\subsection{Ablation and Sensitivity Analysis}

Table~\ref{tab:ablation} examines the contribution of pretraining and
downstream training hyperparameter settings. Removing pretraining reduces Mean
Dice from 0.8686 to 0.0363, indicating that pretrained representation quality
is the dominant factor in downstream performance. Downstream tuning further
improves performance, with background enrichment and BCE-weighted training
achieving the best Mean Dice of 0.9113 and Pooled Dice of 0.9302. Overall,
these results suggest that hyperparameter tuning provides additional gains,
while stronger pretrained representations remain substantially more important
for effective dense feature transfer.

\begin{table}[t]
    \centering
    \caption{Ablation of pretraining and downstream training factors for full-image segmentation. Best and second-best metric values are
shown in \textbf{bold} and \underline{underlined}, respectively.}
    \label{tab:ablation}

    \setlength{\tabcolsep}{2.5pt}
    \renewcommand{\arraystretch}{2.}

    \resizebox{\columnwidth}{!}{
    \begin{tabular}{lcccc}
        \toprule
        \textbf{Setting} &
        \textbf{FG:BG} &
        \textbf{BCE:Dice Weight} &
        \textbf{Mean Dice} &
        \textbf{Pooled Dice} \\
        \midrule

        Without Pretraining &
        1:3 &
        1:1 &
        0.0363 &
        0.0366 \\

        Baseline &
        1:3 &
        1:1 &
        0.8686 &
        0.8959 \\

        Increased Background &
        All BG ($\sim$1:15) &
        1:1 &
        \underline{0.9033} &
        \underline{0.9234} \\

        Dice-Weighted Loss &
        1:3 &
        1:5 &
        0.8030 &
        0.8414 \\

        BCE-Weighted Loss &
        1:3 &
        5:1 &
        0.8899 &
        0.9129 \\

        \textbf{Combined} &
        \textbf{All BG ($\sim$1:15)} &
        \textbf{5:1} &
        \textbf{0.9113} &
        \textbf{0.9302} \\

        \bottomrule
    \end{tabular}
    }
\end{table}

\subsection{Phenotype Quantification}

As a downstream scientific application, we estimate projected leaf area from
the reconstructed full-image segmentations and compare it with manual
measurements. The predicted leaf area shows a consistent linear relationship
with the reference measurements, with a regression slope of 0.993 and a MAPE
of 14.2\%. Given the limited number of available reference measurements, these
results provide preliminary evidence that full-image segmentation can support
downstream quantitative phenotype estimation.

\section{Conclusion}

This work proposes a scalable in-domain self-supervised learning framework for
high-resolution, high-pixel-dimension plant imaging. Pretraining on more than
10 million unlabeled image tiles improves full-image dense segmentation over
natural-image pretrained representations and maintains stronger performance
under limited annotations, cross-view variation, and resolution degradation.
Increasing pretraining exposure further provides consistent gains in dense
feature transfer, while its benefit is smaller for coarse global retrieval.
Preliminary phenotype results also suggest that full-image segmentation can
support downstream quantitative scientific analysis. Overall, these findings
highlight the value of scaling in-domain self-supervised learning for dense
feature analysis of high-pixel-dimension scientific imagery.

\section*{Acknowledgment}

This material by the Orchestrated Platform for Autonomous Laboratories (OPAL) is based upon work supported by the U.S. Department of Energy, Office of Science, through the Office of Biological and Environmental Research Program, under contracts DE-AC02-06CH11357 (ANL); DE-AC02-05CH11231 (LBNL); DE-AC05-00OR22725 (ORNL); and DE-AC05-76RL01830 (PNNL). This research used resources of the Oak Ridge Leadership Computing Facility and the Advanced Plant Phenotyping Laboratory at the Oak Ridge National Laboratory, supported by the Office of Science of the U.S. Department of Energy under Contract No. DE-AC05-00OR22725.

% =========================
% References
% =========================
% Keep references in references.bib.
% Do not manually add \section*{References}; BibTeX will generate it.
\bibliographystyle{IEEEtran}
\bibliography{references}

% Generated by IEEEtran.bst, version: 1.14 (2015/08/26)
\begin{thebibliography}{10}
\providecommand{\url}[1]{#1}
\csname url@samestyle\endcsname
\providecommand{\newblock}{\relax}
\providecommand{\bibinfo}[2]{#2}
\providecommand{\BIBentrySTDinterwordspacing}{\spaceskip=0pt\relax}
\providecommand{\BIBentryALTinterwordstretchfactor}{4}
\providecommand{\BIBentryALTinterwordspacing}{\spaceskip=\fontdimen2\font plus
\BIBentryALTinterwordstretchfactor\fontdimen3\font minus \fontdimen4\font\relax}
\providecommand{\BIBforeignlanguage}[2]{{%
\expandafter\ifx\csname l@#1\endcsname\relax
\typeout{** WARNING: IEEEtran.bst: No hyphenation pattern has been}%
\typeout{** loaded for the language `#1'. Using the pattern for}%
\typeout{** the default language instead.}%
\else
\language=\csname l@#1\endcsname
\fi
#2}}
\providecommand{\BIBdecl}{\relax}
\BIBdecl

\bibitem{remote_sensing_bigdata}
Y.~Li, J.~Ma, and Y.~Zhang, ``Image retrieval from remote sensing big data: A survey,'' \emph{Information Fusion}, vol.~67, pp. 94--115, 2021.

\bibitem{guo2025evaluating}
J.~Guo, S.~Lu, C.~Cui, R.~Deng, T.~Yao, Z.~Tao, Y.~Lin, M.~Lionts, Q.~Liu, J.~Xiong \emph{et~al.}, ``Evaluating cell ai foundation models in kidney pathology with human-in-the-loop enrichment,'' \emph{Communications Medicine}, vol.~5, no.~1, p. 495, 2025.

\bibitem{asign}
J.~Zhu, R.~Deng, T.~Yao, J.~Xiong, C.~Qu, J.~Guo, S.~Lu, M.~Yin, Y.~Wang, S.~Zhao, H.~Yang, and Y.~Huo, ``Asign: An anatomy-aware spatial imputation graphic network for 3d spatial transcriptomics,'' in \emph{2025 IEEE/CVF Conference on Computer Vision and Pattern Recognition (CVPR)}, 2025, pp. 30\,829--30\,838.

\bibitem{plantphenotype1}
R.~Pieruschka and U.~Schurr, ``Plant phenotyping: past, present, and future,'' \emph{Plant Phenomics}, 2019.

\bibitem{plantphenotype2}
N.~Fahlgren, M.~Feldman, M.~A. Gehan, M.~S. Wilson, C.~Shyu, D.~W. Bryant, S.~T. Hill, C.~J. McEntee, S.~N. Warnasooriya, I.~Kumar \emph{et~al.}, ``A versatile phenotyping system and analytics platform reveals diverse temporal responses to water availability in setaria,'' \emph{Molecular plant}, vol.~8, no.~10, pp. 1520--1535, 2015.

\bibitem{ahn2018affinity}
J.~Ahn and S.~Kwak, ``Learning pixel-level semantic affinity with image-level supervision for weakly supervised semantic segmentation,'' in \emph{Proceedings of the IEEE Conference on Computer Vision and Pattern Recognition}, 2018, pp. 4981--4990.

\bibitem{imagenet}
J.~Deng, W.~Dong, R.~Socher, L.-J. Li, K.~Li, and L.~Fei-Fei, ``Imagenet: A large-scale hierarchical image database,'' in \emph{2009 IEEE conference on computer vision and pattern recognition}.\hskip 1em plus 0.5em minus 0.4em\relax Ieee, 2009, pp. 248--255.

\bibitem{simclr}
T.~Chen, S.~Kornblith, M.~Norouzi, and G.~Hinton, ``A simple framework for contrastive learning of visual representations,'' in \emph{International conference on machine learning}.\hskip 1em plus 0.5em minus 0.4em\relax PmLR, 2020, pp. 1597--1607.

\bibitem{moco}
K.~He, H.~Fan, Y.~Wu, S.~Xie, and R.~Girshick, ``Momentum contrast for unsupervised visual representation learning,'' in \emph{2020 IEEE/CVF conference on computer vision and pattern recognition (CVPR)}.\hskip 1em plus 0.5em minus 0.4em\relax IEEE, 2020, pp. 9726--9735.

\bibitem{dino}
M.~Caron, H.~Touvron, I.~Misra, H.~J{\'e}gou, J.~Mairal, P.~Bojanowski, and A.~Joulin, ``Emerging properties in self-supervised vision transformers,'' in \emph{2021 IEEE/CVF international conference on computer vision (ICCV)}.\hskip 1em plus 0.5em minus 0.4em\relax IEEE, 2021, pp. 9630--9640.

\bibitem{beit}
H.~Bao, L.~Dong, S.~Piao, and F.~Wei, ``Beit: Bert pre-training of image transformers,'' \emph{arXiv preprint arXiv:2106.08254}, 2021.

\bibitem{mae}
K.~He, X.~Chen, S.~Xie, Y.~Li, P.~Doll{\'a}r, and R.~Girshick, ``Masked autoencoders are scalable vision learners,'' in \emph{2022 IEEE/CVF conference on computer vision and pattern recognition (CVPR)}.\hskip 1em plus 0.5em minus 0.4em\relax IEEE, 2022, pp. 15\,979--15\,988.

\bibitem{ibot}
J.~Zhou, C.~Wei, H.~Wang, W.~Shen, C.~Xie, A.~Yuille, and T.~Kong, ``ibot: Image bert pre-training with online tokenizer,'' \emph{arXiv preprint arXiv:2111.07832}, 2021.

\bibitem{dinov2}
M.~Oquab, T.~Darcet, T.~Moutakanni, H.~Vo, M.~Szafraniec, V.~Khalidov, P.~Fernandez, D.~Haziza, F.~Massa, A.~El-Nouby \emph{et~al.}, ``Dinov2: Learning robust visual features without supervision,'' \emph{arXiv preprint arXiv:2304.07193}, 2023.

\bibitem{dinov3}
O.~Sim{\'e}oni, H.~V. Vo, M.~Seitzer, F.~Baldassarre, M.~Oquab, C.~Jose, V.~Khalidov, M.~Szafraniec, S.~Yi, M.~Ramamonjisoa \emph{et~al.}, ``Dinov3,'' \emph{arXiv preprint arXiv:2508.10104}, 2025.

\bibitem{lu2025vision}
S.~Lu, J.~Guo, J.~R. Zimmer-Dauphinee, J.~M. Nieusma, X.~Wang, P.~VanValkenburgh, S.~A. Wernke, and Y.~Huo, ``Vision foundation models in remote sensing: A survey,'' \emph{IEEE Geoscience and Remote Sensing Magazine}, vol.~13, no.~3, pp. 190--215, 2025.

\bibitem{vorontsov2023virchow}
E.~Vorontsov, A.~Bozkurt, A.~Casson, G.~Shaikovski, M.~Zelechowski, S.~Liu, K.~Severson, E.~Zimmermann, J.~Hall, N.~Tenenholtz \emph{et~al.}, ``Virchow: A million-slide digital pathology foundation model,'' \emph{arXiv preprint arXiv:2309.07778}, 2023.

\bibitem{szwarcman2025prithvi}
D.~Szwarcman, S.~Roy, P.~Fraccaro, O.~E. G{\'\i}slason, B.~Blumenstiel, R.~Ghosal, P.~H. De~Oliveira, J.~L. de~Sousa~Almeida, R.~Sedona, Y.~Kang \emph{et~al.}, ``Prithvi-eo-2.0: A versatile multi-temporal foundation model for earth observation applications,'' \emph{IEEE Transactions on Geoscience and Remote Sensing}, 2025.

\bibitem{cong2022satmae}
Y.~Cong, S.~Khanna, C.~Meng, P.~Liu, E.~Rozi, Y.~He, M.~Burke, D.~Lobell, and S.~Ermon, ``Satmae: Pre-training transformers for temporal and multi-spectral satellite imagery,'' \emph{Advances in Neural Information Processing Systems}, vol.~35, pp. 197--211, 2022.

\bibitem{reed2023scale}
C.~J. Reed, R.~Gupta, S.~Li, S.~Brockman, C.~Funk, B.~Clipp, K.~Keutzer, S.~Candido, M.~Uyttendaele, and T.~Darrell, ``Scale-mae: A scale-aware masked autoencoder for multiscale geospatial representation learning,'' in \emph{2023 IEEE/CVF International Conference on Computer Vision (ICCV)}.\hskip 1em plus 0.5em minus 0.4em\relax IEEE, 2023, pp. 4065--4076.

\bibitem{11196959}
J.~Guo, J.~R. Zimmer-Dauphinee, J.~M. Nieusma, S.~Lu, Q.~Liu, R.~Deng, C.~Cui, J.~Yue, Y.~Lin, T.~Yao, J.~Xiong, J.~Zhu, C.~Qu, Y.~Yang, M.~Wilkes, X.~Wang, P.~VanValkenburgh, S.~A. Wernke, and Y.~Huo, ``Deepandes: A self-supervised vision foundation model for multispectral remote sensing imagery of the andes,'' \emph{IEEE Journal of Selected Topics in Applied Earth Observations and Remote Sensing}, vol.~18, pp. 26\,983--26\,999, 2025.

\bibitem{wang2025orbit}
X.~Wang, J.-Y. Choi, T.~Kurihaya, I.~Lyngaas, H.-J. Yoon, X.~Xiao, D.~Pugmire, M.~Fan, N.~M. Nafi, A.~Tsaris \emph{et~al.}, ``Orbit-2: Scaling exascale vision foundation models for weather and climate downscaling,'' in \emph{proceedings of the International Conference for high performance computing, networking, storage and analysis}, 2025, pp. 86--98.

\bibitem{milligan2026vision}
J.~Milligan, A.~Seethepalli, A.~Tsaris, X.~Wang, L.~York, and J.~H. Lagergren, ``Vision transformers enable advanced plant phenotyping in controlled environments,'' \emph{bioRxiv}, pp. 2026--09, 2026.

\end{thebibliography}

\end{document}